\documentclass{article}
\usepackage{ijcai16}

\usepackage{times}
\usepackage{graphicx}
\usepackage{caption}
\usepackage{subcaption}
\usepackage{amsmath}
\usepackage{mathrsfs}
\usepackage{cases}
\usepackage{float}
\usepackage{array}
\usepackage{multirow}

\numberwithin{equation}{section}

\newcommand{\specialcell}[2][c]{%
  \begin{tabular}[#1]{@{}c@{}}#2\end{tabular}}

\title{Deep Convolutional Neural Network for 6-DOF Image Localization}
\author{Daoyuan Jia ~~~~ Yongchi Su ~~~~ Chunping Li\\
\\ 
 Tsinghua University, Beijing, China  \\
 jiady14, suyc13, cli @mails.tsinghua.edu.cn}

\begin{document}

\maketitle

\begin{abstract}
Wee present an accurate and robust method for six degree of freedom image localization. There are two key-points of our method, 1). automatic immense photo synthesis and labeling from point cloud model and, 2). pose estimation with deep convolutional neural networks (ConvNets) regression. Our model can directly regresses 6-DOF camera poses from images, accurately describing where and how it was captured.  We achieved an accuracy within 1 meters and 1 degree on our out-door dataset, which covers about $20,000m^{2}$ on our school campus. Unlike previous point cloud registration solutions, our model supports low resolution images (i.e. $224\times224$ in our settings), and is tiny in size when finished training. Moreover, in pose estimation, our model uses $O(1)$ time \& space complexity as trainset grows. We will show the importance to localization using hundreds of thousands of generated and self-labeled "photos" came from a short video. We will show our model's robustness despite of illumination and seasonal variances, which usually fails methods that leverage image feature descriptors like SIFT. Furthermore, we will show the ability of transfer our model trained on one scene to another, and the gains in accuracy and efficiency. 
\end{abstract}

\section{Introduction}

Image localization, or camera relocalization, describes a problem of approximating the position and orientation of the camera when the query image was taken. This problem inspires and connects to many cutting-edge applications in our life, such as self-driving automobiles, unmanned aerial vehicles (UAVs), virtual reality (VR) and etc.

For years, researchers came to many solutions based on visual descriptors of images, such as SIFT \cite{lowe2004distinctive} or SURF \cite{bay2006surf}. We categorize these methods into three main genres, {\it Descriptors Matching},  {\it Dynamic Reconstruction}, and {\it Point Cloud}, as well as combinations and optimizations of them. These solutions usually demand so much computation / storage resource that invalidates their applications on mobile devices, or restricted to small scene usage, or require Internet and server-side participation, which is impractical in some daily scenarios. Our method distinguish from them as we provide a complete end-to-end solution with deep convolutional neural network (ConvNet), which directly takes query images as input and regress camera position and orientation from them. 

One of the main problem in exploiting a deep ConvNet is the sparsity of train samples relative to network's depth and vast number of parameters. Several thousands of photos are sufficient to reconstruct a 3D model, but far from for training a ConvNet to regress camera pose, and would almost certainly lead to over-fitting. We coped this problem by integrating Point Cloud solution. We reconstructed the 3D point cloud model of the scene, and from which we can generate hundreds of thousands of "photos" [增加了信息], all of which contains accurate 6-DOF coordinates relative the model. This also solved another big problem in image localization evaluations we met, which We call it {\it Groundtruth Dilemma}. It is because that the 6-DOF ground truth of test-set and train-set is not easily available. Datasets like Quad \cite{disco2013pami} and Aachen \cite{middelberg2014scalable} were collected with differential GPS to obtain positions with a precision of $10cm$, and only a handful of test images come with accurate orientation angles. We captured videos in our school campus and split them into frames. In total we created 3 [**] scenes, and reconstructed 3D points cloud model for each of them. From each model we generated more than 100,000 accurately labeled "photos", of various positions and locations, to feed our deep ConvNets. We also took advantage of some shaders to augment our dataset, so that the trained deep ConvNets can tolerate weather and light variances. 

We modified and experimented on CaffeNet \cite{NIPS2012_4824}, GoogLeNet \cite{szegedy2015going} and VGG \cite{simonyan2014very}. Due to the depth of some ConvNets, we modified layer structures and adopted different train policy. We also found it is very helpful to use pre-trained models, which has already been trained on other giant dataset like ImageNet \cite{ILSVRC15}. Pre-trained model serves as a better initialization, accelerated our training process and produced better converged result. We finally reached an average accuracy of 1 meters in position and 0.8 degree in orientation. We extended our experiments to photos in different illumination, season conditions, and found our model is robust to these changes. We also transfered our trained model of one scene to another, and found that training on new scene takes less epochs to converge. 

Our contributions are in the following areas. 
\begin{itemize}
\item we are the first to use synthesis method to generate immense number of accurately 6-DOF labeled photos from point cloud and shaders, to build and further extend dataset for camera relocalization. This improves the pose estimation accuracy of deep ConvNets and robustness to various light, weather condition;
\item we modified and applied three state of art deep ConvNet architectures from image classification to pose regression and evaluated their result. 
\end{itemize}

The organization of the article is as follow: related work will be covered in section 2; our pipeline from input scene video to network training is described in section 3; section 4 provides experimental evaluation and analysis; we summarize our work in section 6.

\section{Related work}

We categorize current visual descriptors based image localization methods into three main genres,
{\it Descriptors Matching}, methods used by \cite{li2010location} and \cite{li2009landmark} aimed at estimating a camera pose by matching descriptors to a known scene. This genre of methods solve localization as a process of image retrieval by find images in the databases that shared most visual descriptors. If images in the database are labeled with GPS tags, the location of the query one can be answered by averaging existed ones. For instance, \cite{schindler2007city} and \cite{chen2011city} seek to localize urban images by matching images in Google street-view. Accuracy of methods in this genre is usually coarse to tens or hundreds of meters in outdoor scene, and seldom supports 6-DOF localization. Also, as the dataset expands, the query time increase accordingly. {\it Dynamic Reconstruction}, \cite{klein2007parallel} and \cite{castle2008video} utilized simultaneous localization and mapping (SLAM) to concurrently reconstruct a 3D model and estimate the camera pose from extracted visual descriptors. As the 3D maps grows larger, the mapping part of the SLAM becomes more expensive to run on mobile devices. {\it Point Cloud} methods \cite{sattler2011fast} and \cite{donoser2014discriminative}, pre-build a model, using methods like Structure-from-Motion (SfM) \cite{agarwal2009building} to reconstruct a scene composed with tens of millions of 3D points. This process requires intensive computation to extract millions SIFT features from thousands of photos and find concurrences among them. Moreover, in the query time, it consumes loads of memory and computing resource to establish 2D-3D correspondences between scene points and visual features points from query images. \cite{li2012worldwide} and \cite{sattler2012improving} introduced active 3D-2D correspondences match to improve registration rate and refine the result. \cite{middelberg2014scalable} combined SLAM and point cloud solutions together. They incrementally apply local relative movements measured by SLAM algorithm onto global position previously calculated by point cloud on the server to obtain current global position. The decoupling of computation of relative movement and global position reduces computation on local devices (i.e. mobile devices), but the computation on the server is still heavy and this pipeline requires network participation at the very beginning. All these three genres of methods usually require resources exceeding computation and memory capacities of modern mobile devices, and thus invalidated their applications. 

Apart from this, some researchers focused on regression methods on RGB-D images. \cite{shotton2013scene} used a regression forest to infer correspondences between each depth image pixels and points in the scene, which was constructed with RGB-D input images. Their method amended the problem of too few 2D-3D matches (in-liners correspondences) in point cloud registration method, as they allow densely or sparsely on points sampling in depth images. \cite{krull20146} introduced a two-way procedure to integrate the random forest regression the distribution generation to improve the generation ability to deal with occlusion. Their accuracy of 6-DOF pose estimation exceeded state-of-art methods on certain public datasets. Methods of this type are usually conducted in small indoor scene and require depth information in training. In contrast, our method only use a monocular camera, available on almost every cell phone, to collect necessary data, and evaluations indicate our method works well on bigger scene. 

Convolutional Neural Network (ConvNet) has been proved to be potent in recognition of many fields, i.e. GoogLeNet and VGG excelled in Ilsvrc in these two years and are well applied to other recognition tasks. GoogLeNet \cite{szegedy2015going} introduced "Inception", a unit composed of multiple-sized convolutional / pooling layers. It reduced parameter numbers and enabled greater depth. VGG \cite{simonyan2014very} exploited very small convolution filters. They used filters of size $1\times1$ through $3\times3$, and stacked several convolutional layers in conjunction. Stacking conv-layers of small filter is equivalent to a conv-layer of bigger filter. It would provide better generation ability, and reduced parameters than single conv-pooling layer with big filter. 

\cite{kendall2015posenet} follows \cite{shotton2013scene}, and used a modified GoogLeNet to regress pose directly. They added another layer of size 2048 before final regression layer, and took advantage of model pre-trained on ImageNet and Flicker datasets. The transfer learning model accelerated the training on camera relocalization, and decreased the number of training samples to use. They reached an accuracy of about $3m$ in position and $5deg$ in orientation. We follows the "transfer" idea and extended choice of deep ConvNets to CaffeNet, GoogLeNet and VGG. We applied layer-wise training and fine-tuned with different pre-trained models to reach the best performance. Moreover, we detailed our method of photo synthesis and dataset augmentation in different condition.

\section{Estimate camera pose with ConvNets}
In this section we will present our way of  1). automatic generate loads of labeled data and, 2). modeling the question and ConvNet configurations of CaffeNet, GoogLeNet, and VGG.

\subsection{Automatic data generation and labeling}

\begin{figure*}
    \centering
    \begin{subfigure}[b]{0.49\textwidth}
        \includegraphics[width=0.54\textwidth]{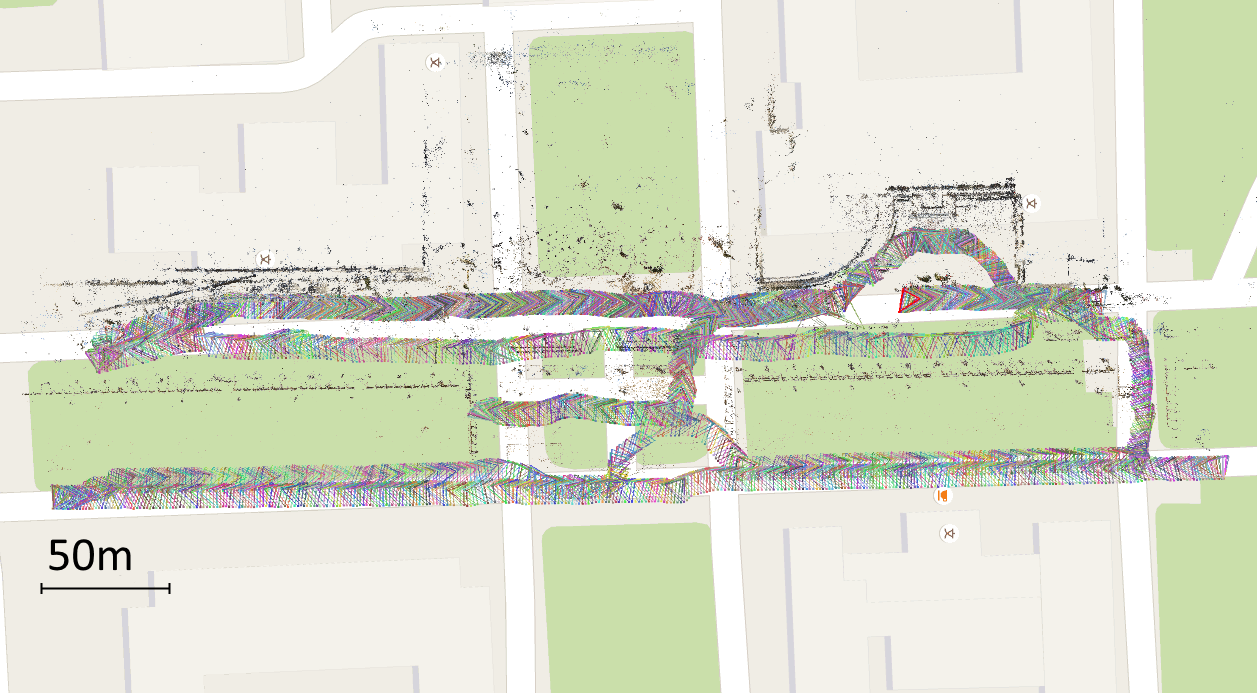}
        \includegraphics[width=0.44\textwidth]{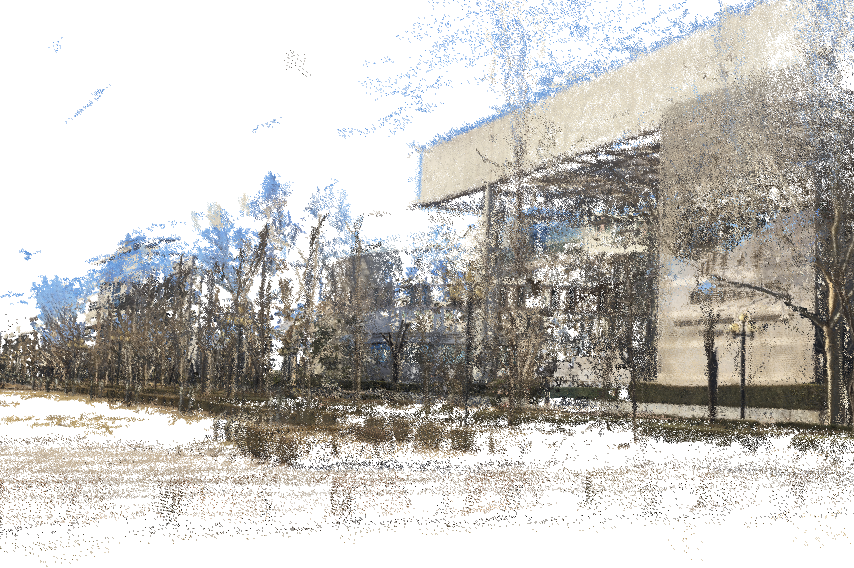}
        \caption{Law school}
        \label{fig:law}
    \end{subfigure}
    ~ 
    \begin{subfigure}[b]{0.49\textwidth}
        \includegraphics[width=0.49\textwidth]{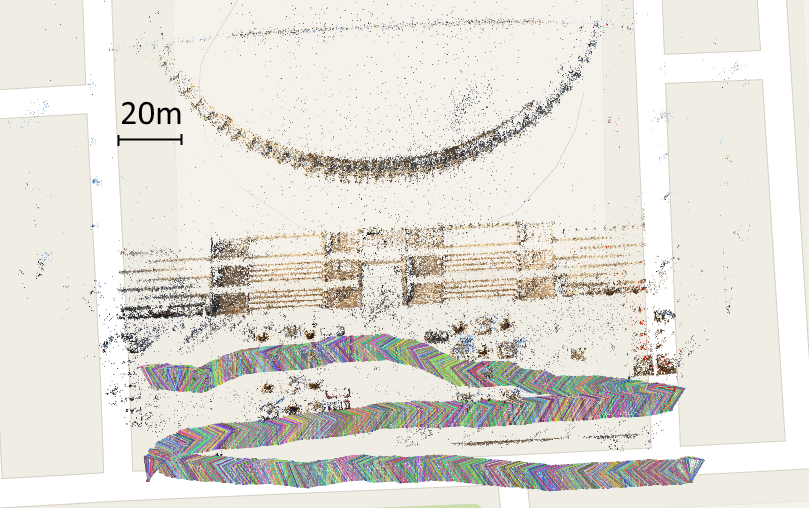}
        \includegraphics[width=0.49\textwidth]{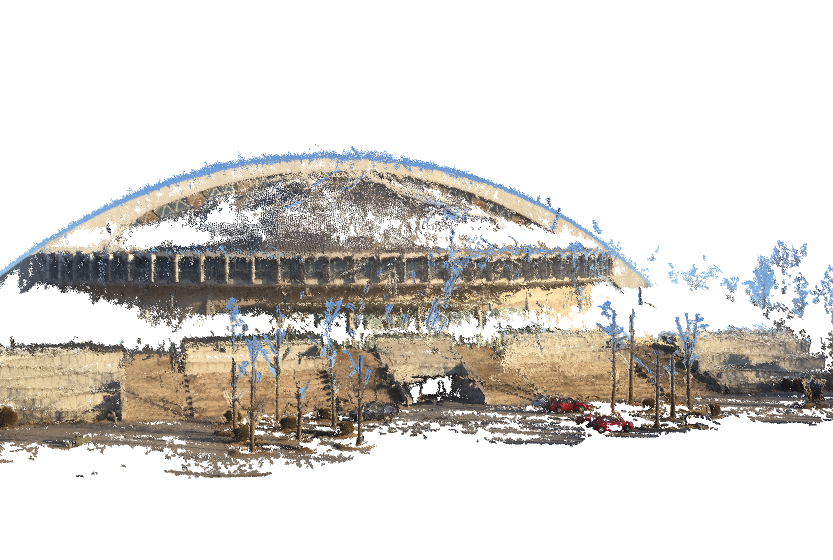}
        \caption{School stadium}
        \label{fig:stad}
    \end{subfigure}
    \caption{Planar view showing the video frames positions of dataset and corresponding dense point cloud model of scene law school and school stadium.}
    \label{fig:bird_view_map}
\end{figure*}

In previous camera relocalization regression solutions, photos were regarded as individual. We think it is beneficial to correlation information of photos to train a regressors.Based on this intuition, we designed a way to generate labeled photos does not originally exist in dataset.
\begin{figure}
    \centering
    \begin{subfigure}[b]{0.27\textwidth}
        \includegraphics[width=\textwidth]{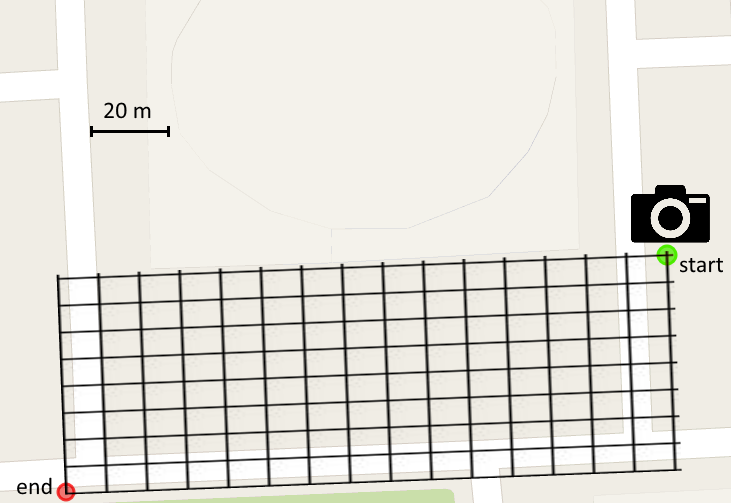}
        \caption{Camera translation grid}
        \label{fig:camera_grid}
    \end{subfigure}
    ~ 
    \begin{subfigure}[b]{0.2\textwidth}
        \includegraphics[width=\textwidth]{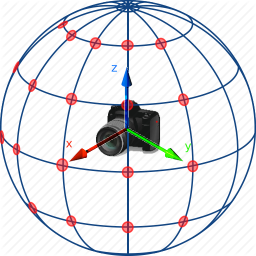}
        \caption{Camera rotation sphere}
        \label{fig:camera_sphere}
    \end{subfigure}
    ~
	\begin{subfigure}[b]{0.49\textwidth}
		\includegraphics[width=0.24\textwidth]{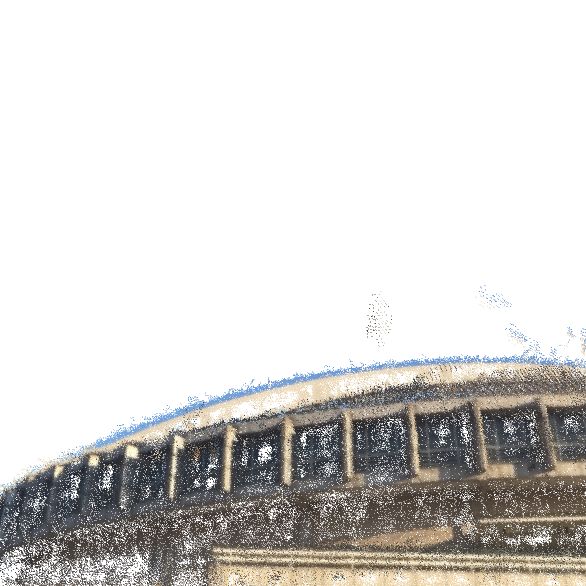}
		\includegraphics[width=0.24\textwidth]{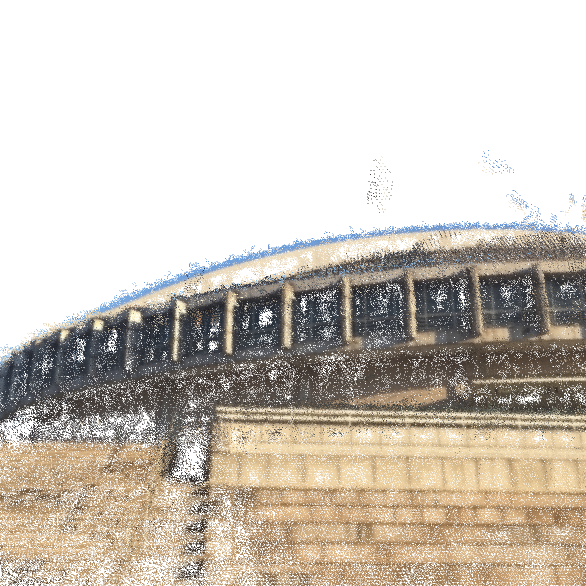}
		\includegraphics[width=0.24\textwidth]{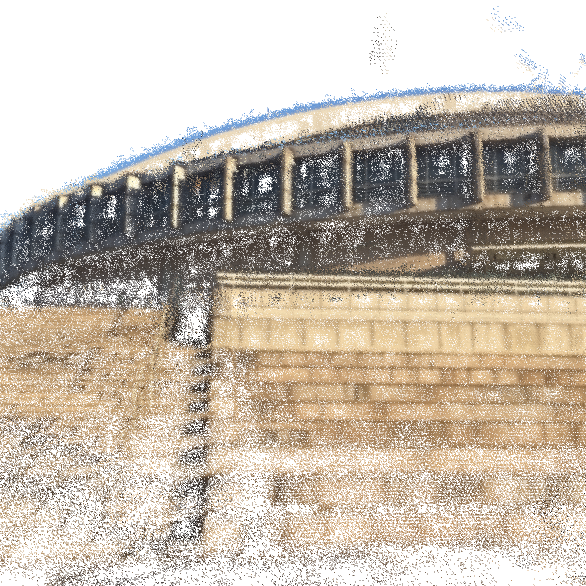}
		\includegraphics[width=0.24\textwidth]{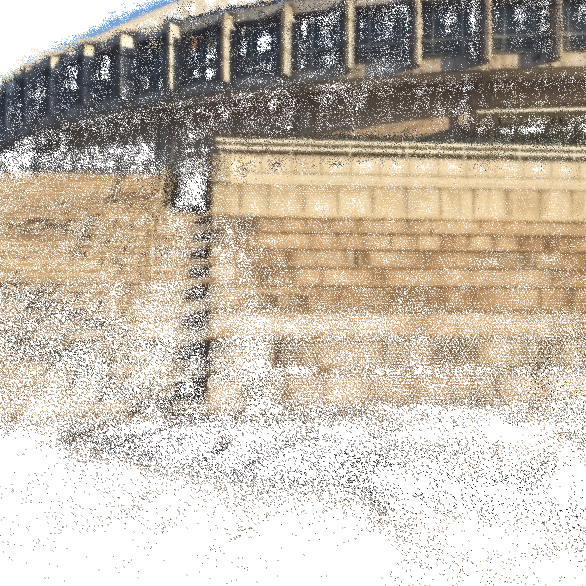}
		\caption{Synthesized school stadium photo samples}
		\label{fig:gen_samples}
	\end{subfigure}
	~
	\begin{subfigure}[b]{0.49\textwidth}
		\includegraphics[width=0.24\textwidth]{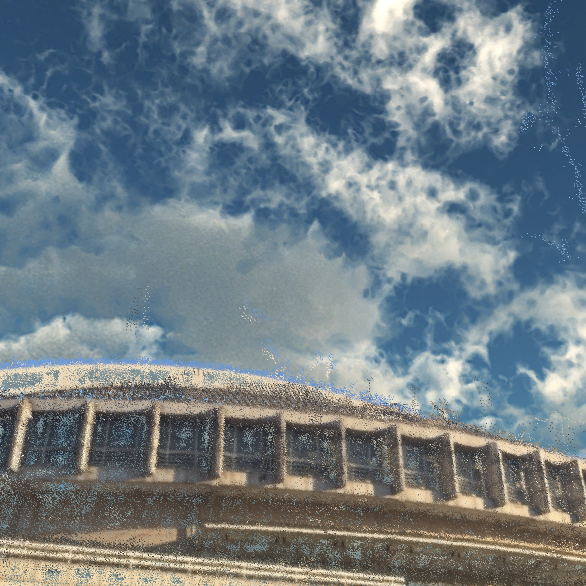}
		\includegraphics[width=0.24\textwidth]{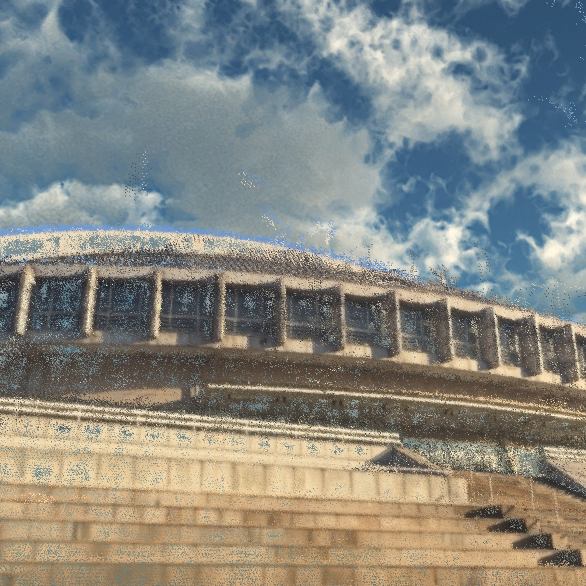}
		\includegraphics[width=0.24\textwidth]{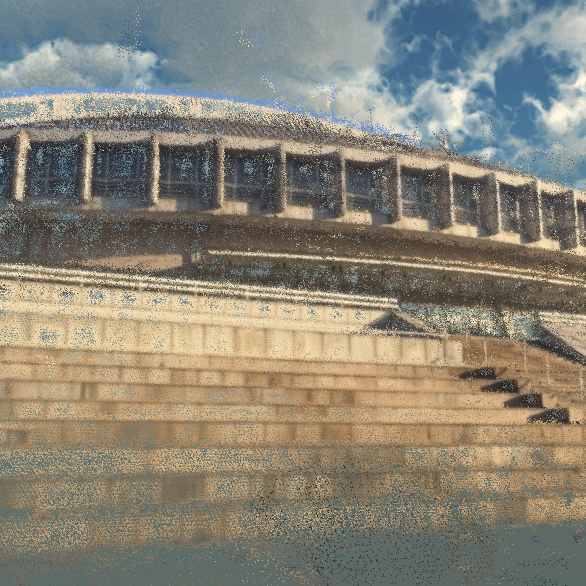}
		\includegraphics[width=0.24\textwidth]{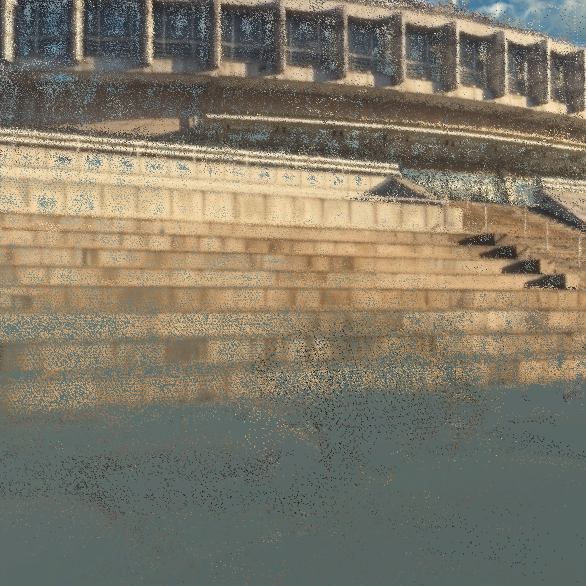}
		\caption{Synthesized noon school stadium with sky-box}
		\label{fig:gen_samples_noon}
	\end{subfigure}
	~
	\begin{subfigure}[b]{0.49\textwidth}
		\includegraphics[width=0.24\textwidth]{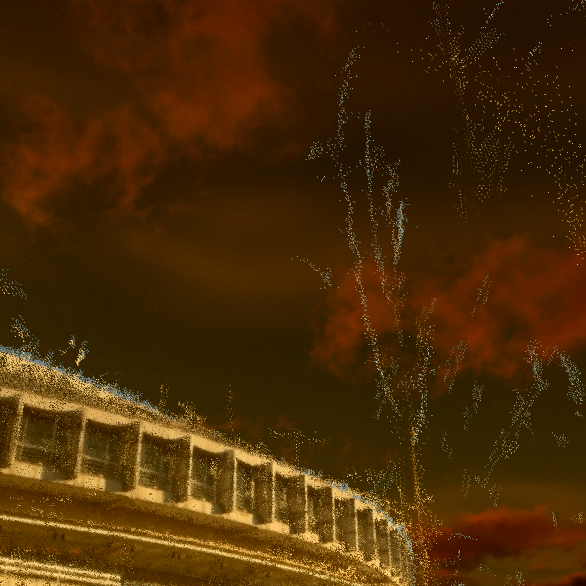}
		\includegraphics[width=0.24\textwidth]{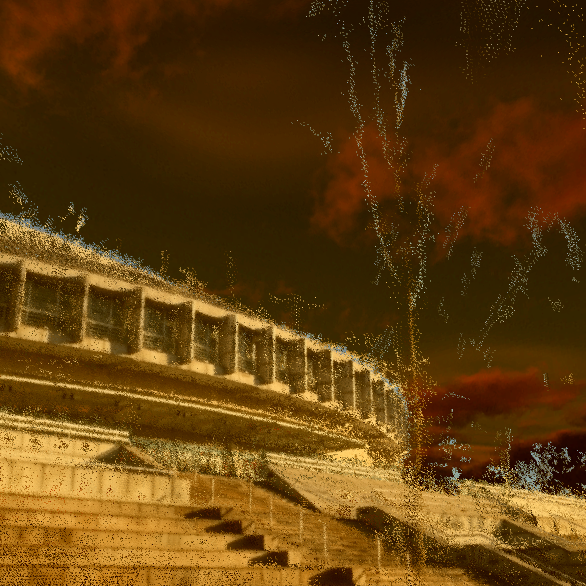}
		\includegraphics[width=0.24\textwidth]{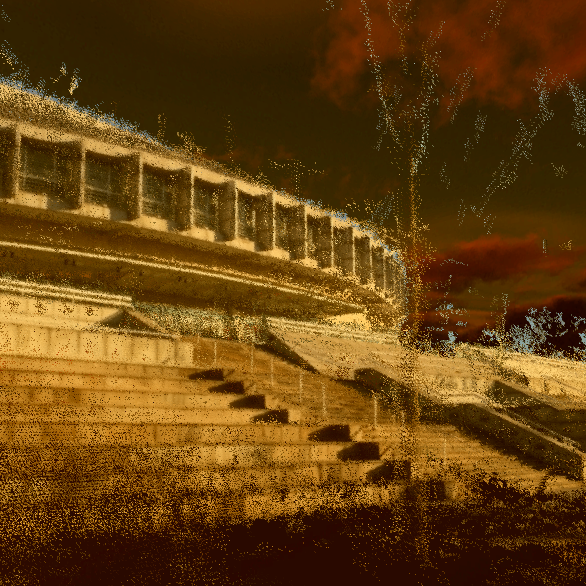}
		\includegraphics[width=0.24\textwidth]{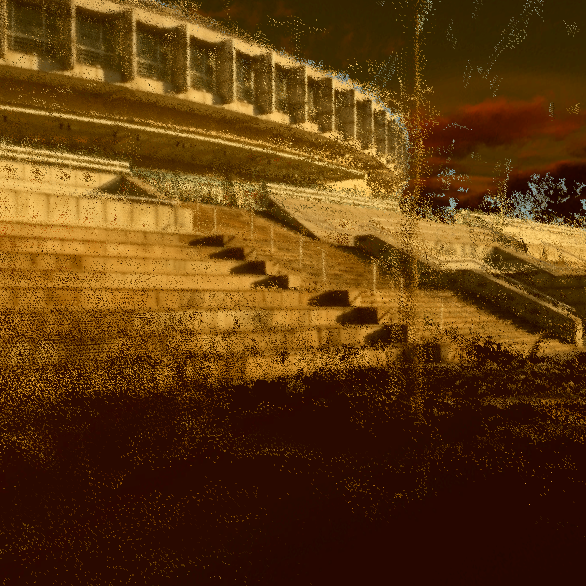}
		\caption{Synthesized dusk school stadium with sky-box and shader}
		\label{fig:gen_samples_dust}
	\end{subfigure}
    \caption{Illustration of camera position/orientation update policy. Fig.\ref{fig:camera_grid} shows the track of virtual camera movement in scene school stadium. From green dot to red dot, camera take photos of different orientations at each cross dot. Fig.\ref{fig:camera_sphere} shows the camera rotation. Camera changes its orientation to each red dot, and take photos. In fig.\ref{fig:gen_samples}-\ref{fig:gen_samples_dust}, samples of synthesized photos in school stadium scene are shown. Photos were rendered with sky-box and shader to augment reality.}
    \label{fig:pose_update_policy}
\end{figure}
We take input of video, and split it into frames at a frequency of 3Hz. Then, frames are used to reconstruct dense point cloud model of the scene using SfM. In implementation, we used VisualSfM \cite{wu2013towards} and PVMS \cite{furukawa2010accurate}, which take advantage of GPU acceleration and multicore bundle adjustment, to fast build the model. In ~fig.\ref{fig:bird_view_map} we illustrated the bird view of our frames and corresponding densely reconstructed model.

Dense point cloud model was imported into Unity to synthesize photos. For simplicity, in Unity we used the coordinate system generated by VisualSfM. This will not bring trouble because synthesized photos and origin frames shared the same coordinate system, and it does not matter where the origin was set. Therefore, we can create a virtual camera in Unity, recalibrate and synchronize its instincts with our real cell phone. After that, we constantly 1). take a photo with the virtual camera, 2). take down it pose, 3). update it to a new position / orientation.

The position/orientation update policy is illustrated in fig.\ref{fig:pose_update_policy}. The track of virtual camera translation is a grid. At each cross dot, it takes several photos with various orientation, and record the corresponding 6-DOF coordinates. Furthermore, we can utilize different shader/sky-box. Fig.\ref{fig:gen_samples}-\ref{fig:gen_samples_dust} in turn shows the synthesized photos with pure model, synthesized noon photos rendered sunny sky-box, and synthesized dusk photos rendered with evening sky-box and shader. This method enables us to simulate photos of differed time-of-the-day, weather condition and etc.

The synthesis method is meaningful not only because its ability to automatic generate tons of labeled photos at a short time, but lies in the fact that \textbf{it fully utilized information across original video frames}. That is, a synthesized photo may includes details from separated frames, i.e. frame 1 and frame 1000, determined by the principle of SfM. \textbf{Using these massive fabricated labeled data will prevent pose regression deep ConvNet from overfitting and help ConvNets learn even nuances of spatial variances}.   

\subsection{Modeling the question}
We represent image $\boldsymbol{I}$ as a matrix of pixels, and its pose $\boldsymbol{P}$ as a 6 dimension vector, represents as a combination of position $p$ and orientation $r$. 
\begin{equation}
\boldsymbol{P}=[p; r]
\end{equation}
Position $p$ is a 3 dimension vector, and orientation $r$ is composed with pitch, yaw, and roll. We globally use $meter$ and $degree$ as their respective unit. Our target is to approximate a function $\mathscr{F}$ using ConvNet that takes $\boldsymbol{I}$ as input and produce $\boldsymbol{P}$. Therefore, we want to optimize the an objective loss function like:
\begin{equation}
\mathscr{L}(\boldsymbol{I})= ||\boldsymbol{\mathscr{F}(I) - P}||_2 
\end{equation}
Since we use a deep ConvNet, we can use an Euclidean loss function taking form as: 
\begin{equation}
\mathscr{L^*}(\boldsymbol{I})= ||\boldsymbol{w}^T\boldsymbol{\hat{P}} - \boldsymbol{w}^T\boldsymbol{P}||_2,
\end{equation}
where $\boldsymbol{\hat{P}}$ denotes the predicted pose vector, and $\boldsymbol{w}$ is the weight factor to balance the importance of position and orientation error. By default, we use $\boldsymbol{w=1}$. It means that we regard position error of 1 meter as the same as orientation error of 1 degree. However, if we can tolerate greater orientation error, greater values can be applied to corresponding dimension of $\boldsymbol{w}$.

\subsection{ConvNet Configurations}

For complete of comparison, we conducted our experiments on several state-of-art deep ConvNets, specifically caffeNet, GoogLeNet and VGG16 as our pose regression network. If only counting layers with trainable parameters, caffeNet has 8 layers, and composed with traditional convolution-pooling structure. GoogLeNet has 22 layers, but is organized slightly different. It introduced "inception", a module that consists of horizontal-stacked, various-sized convolution layers. In GoogLeNet, There are 6 "inception" module chained in a row, together with intermediate loss layers to amplify gradient signal. VGG16 differs from both of them in that it uses very small receptive field of size $3x3$ through out the whole net, and it stacks several convolutional layers in conjunction to approximate a conv-layer of greater receptive field. This makes decision function more discriminative and reduced trainable parameters.

All three ConvNets takes image of size $224\times224$ by default. Therefore, we rescaled and cropped images in dataset to this resolution. We applied center crop policy to keep the field-of-view, and disabled data mirror, since mirroring image would confuse the ConvNet by creating "photos-pose" pair that should not exist, i.e. in symmetry some scenes. Apart from this, we found it is important to minus mean value in data layer for final performance. 

We made changes to the network settings in a way similar to \cite{kendall2015posenet} but with some differences:
\begin{itemize}
\item we change the final feature layer size to 6, and increase it learning rate amplifier;
\item we add another full connection layer of right before final feature layer, and increase it learning rate amplifier;
\item for VGG16, we added middle loss layer to amplify gradient signal;
\end{itemize}

\subsection{ConvNet Training}
We utilized \textit{Caffe} \cite{jia2014caffe} to implement our ConvNet training. All three ConvNets were trained with stochastic gradient decent solver, with the same train epochs (300), step length of learning rate decrease (30 epochs) and batch size (90), on a NVidia GTX 980 graphic card. It is necessary to note here that 4GB graphic card memory is limited for deep ConvNets like GoogLeNet and VGG using a big batch. Therefore, we used mini-batch to accumulate gradient over several batches before back-propagation. Instead of training ConvNets from scratch, we exploited pre-trained models, which has been trained to stable on ImageNet dataset. Some varied training strategies were applied on three ConvNets according their instinct characteristics to achieve their best performance. 
\begin{itemize}
\item \textbf{CaffeNet}: Base learning rate $10^-5$, reducing 90\% every step;
\item \textbf{GoogLeNet}: Base learning rate $10^-6$, reducing 80\% every step;
\item \textbf{VGG16}: Base learning rate $10^-6$, reducing 10\%every step.
\end{itemize}

\section{Dataset}

For this paper, we release an outdoor campus localization dataset, with 2 scenes. This dataset contains the original video file we collected with iPhone6 cell phone, video frames and dense point cloud model we generated. We synthesized immense photos with different settings, and included them in the dataset, too. 

This dataset is slightly unusual as it contains two testsets for each scene, namely testsetA and testsetB. TestsetA is composed with real photos, the origin video frames, while testsetB consists of synthesized photos, generated from dense point cloud. Note that testsetA has previously been separated from origin video frames before 3D reconstruction, and registered to point cloud afterwards to obtain their 6-DOF label as ground-truth. This guarantees that there is no intersection between trainset and testsets. We refer frames that were used to reconstruct 3D model as "train frames". The remaining synthesized photos are used as trainset. Train frames can be included into trainset as an augmentation, but is not required, since generated photos outnumber train frames by 2 magnitudes. We illustrated our dataset detail in table~\ref{tbl:dataset_table}.
\begin{table}[H]
    \centering
    \begin{tabular}{   m{12mm} m{8mm} c c m{8mm} m{8mm} } 
    	Scene 			& Area ($m^2$)	& TestsetA 	& TestsetB 	& \#~Train frames & \#~Gen. photo\\
    	\hline\hline
    	school stadium	& 6,000		& 150		& 19k		& 1,200	  		 & 109k \\
    	\hline
    \end{tabular}
    \caption{Dataset details.}
    \label{tbl:dataset_table}
\end{table}

\begin{table*}[!htbp]
    \centering
    \begin{tabular}{l | c | c c c c c} 
    	Scene 						& Testset  & PoseNet* 				& CaffePose 					& VGGPose 		& GoogLePose		& GoogLePose$^\dagger$\\
    	\hline\hline
    	\multirow{2}*{School stadium}	& TestsetA 		
    		& \specialcell{3.03m, $8.15^{\circ}$\\3.77m, $11.0^{\circ}$} 
    		& 4.32m, $8.60^{\circ}$
    		& \specialcell{1.72m, $1.67^{\circ}$\\ - } 
    		& \specialcell{1.54m, $0.92^{\circ}$\\2.25m, $1.59^{\circ}$}
    		& 1.01m, $0.98^{\circ}$\\
    	& TestsetB 		
    		& \specialcell{7.33m, $13.0^{\circ}$\\8.36m, $14.2^{\circ}$}  
    		& 2.32m, $4.51^{\circ}$ 	
    		& 1.17m, $0.75^{\circ}$
    		& \specialcell{0.91m, $0.39^{\circ}$\\1.01m, $1.44^{\circ}$}
    		& 0.93m, $0.42^{\circ}$\\
    	\hline
    \end{tabular}
    \caption{6-DOF regression result.}
    \label{tbl:result_table}
\end{table*}
\section{Experiments}

We will show in this section experimental evaluation of our method pipeline. For readability, we refers \textit{pipeline} used by \cite{kendall2015posenet} as "PoseNet*", whose training data directly came from video frames and coordinates label from SfM result, and regressed result using their modified version of GoogLeNet. Therefore, we only use train frames and corresponding 7 dimension labels (xyz \& rotation quaternion) of our scenes to train a PoseNet* model. 

In contrast, we refers our modification of deep ConvNets, CaffeNet, VGG16 and GoogLeNet as CaffePose, VGGPose, and GoogLePose respectively in the following experimental sections. We train these ConvNets with automatic generated photos and labels proposed sect. 3.1., along with necessary training methods we mentioned before. The dagger symbol ($\dagger$) is used to denote training the ConvNet additionally using train frames. i.e. GoogLePose$^\dagger$.

Table \ref{tbl:result_table} lists the pose estimation accuracy in terms of position and orientation error. We shows that our method is able to accurately localize images on evaluation scenes. In scenes like School stadium, our proposed pipeline produced much more accurate result than PoseNet, improves orientation estimation accuracy to near 10 times and position estimation to 2 times.

\bibliographystyle{named}
\bibliography{ijcai16}

\end{document}